\documentclass[10pt,twocolumn,letterpaper]{article}

\usepackage[pagenumbers]{cvpr} 

\definecolor{cvprblue}{rgb}{0.21,0.49,0.74}
\usepackage[pagebackref,breaklinks,colorlinks,allcolors=cvprblue]{hyperref}

\usepackage{multirow}
\usepackage{xcolor}
\usepackage{colortbl}
\usepackage{bm}
\usepackage{amsmath}
\usepackage{amssymb}
\usepackage{mathtools}
\usepackage{algorithm}
\usepackage{pifont}
\usepackage{amsfonts}
\usepackage{algpseudocode}
\DeclarePairedDelimiter\ceil{\lceil}{\rceil}

\newif\ifdraft
\drafttrue

\renewcommand{\ell}{y}

\definecolor{orange}{rgb}{1,0.5,0}
\definecolor{gr}{rgb}{0,0.65,0}
\definecolor{mygray}{gray}{0.95}

\ifdraft
 \newcommand{\RS}[1]{{\color{red}{\bf RS: #1}}}
 
 \newcommand{\PMN}[1]{{\color{orange}{\bf PMN: #1}}}

\else
 \renewcommand{\sout}[1]{}
 \newcommand{\RS}[1]{{\color{red}{}}}
 
 \newcommand{\PMN}[1]{{\color{red}{}}}
 
\fi

\DeclareMathOperator{\EX}{\mathbb{E}}
\DeclareMathOperator*{\argmin}{arg\,min}

\newcommand{\real}{\mathbb{R}}

\newcommand{\x}{\mathbf{x}}

\newcommand{\J}{\mathbf{J}}

\newcommand{\f}{\mathbf{f}}
\newcommand{\e}{\mathbf{e}}
\newcommand{\g}{\mathbf{g}}

\newcommand{\bv}{\mathbf{v}}

\newcommand{\NL}{{NL-Invs}}

\algnewcommand{\LineComment}[1]{\State \(\triangleright\) #1}

\newcommand{\framework}{\textsc{NCIS}\xspace} 

\title{Non-Linear Outlier Synthesis for Out-of-Distribution Detection}

\author{
Lars Doorenbos
\qquad
Raphael Sznitman 
\qquad
Pablo Márquez-Neila
\vspace{0.5em}
\\
University of Bern, Bern, Switzerland\\
{\tt\small \{lars.doorenbos,raphael.sznitman,pablo.marquez\}@unibe.ch}
}

\begin{document}
\maketitle

\begin{abstract}
  The reliability of supervised classifiers is severely hampered by their limitations in dealing with unexpected inputs, leading to great interest in out-of-distribution (OOD) detection. Recently, OOD detectors trained on synthetic outliers, especially those generated by large diffusion models, have shown promising results in defining robust OOD decision boundaries. Building on this progress, we present \framework, which enhances the quality of synthetic outliers by operating directly in the diffusion's model embedding space rather than combining disjoint models as in previous work and by modeling class-conditional manifolds with a conditional volume-preserving network for more expressive characterization of the training distribution. We demonstrate that these improvements yield new state-of-the-art OOD detection results on standard ImageNet100 and CIFAR100 benchmarks and provide insights into the importance of data pre-processing and other key design choices. We make our code available at \url{https://github.com/LarsDoorenbos/NCIS}.
\end{abstract}
    
\section{Introduction}
\label{sec:intro}

Modern deep learning classifiers can classify unseen images into thousands of classes when trained on sufficiently broad datasets. However, unexpected samples from unseen classes will also be confidently assigned to one of the training classes~\cite{hendrycks2016baseline}. In most cases, model outputs do not provide information about the reliability of a prediction, leading to silent failures that undermine the trustworthiness of these systems. This has led to significant research on detecting and filtering out these unexpected samples, a subject area known as out-of-distribution (OOD) detection. Specifically, OOD detection aims to enhance the reliability of downstream systems by identifying and removing samples that fall outside the known training distribution. 
\begin{figure}
    \centering
    \includegraphics[width=\linewidth]{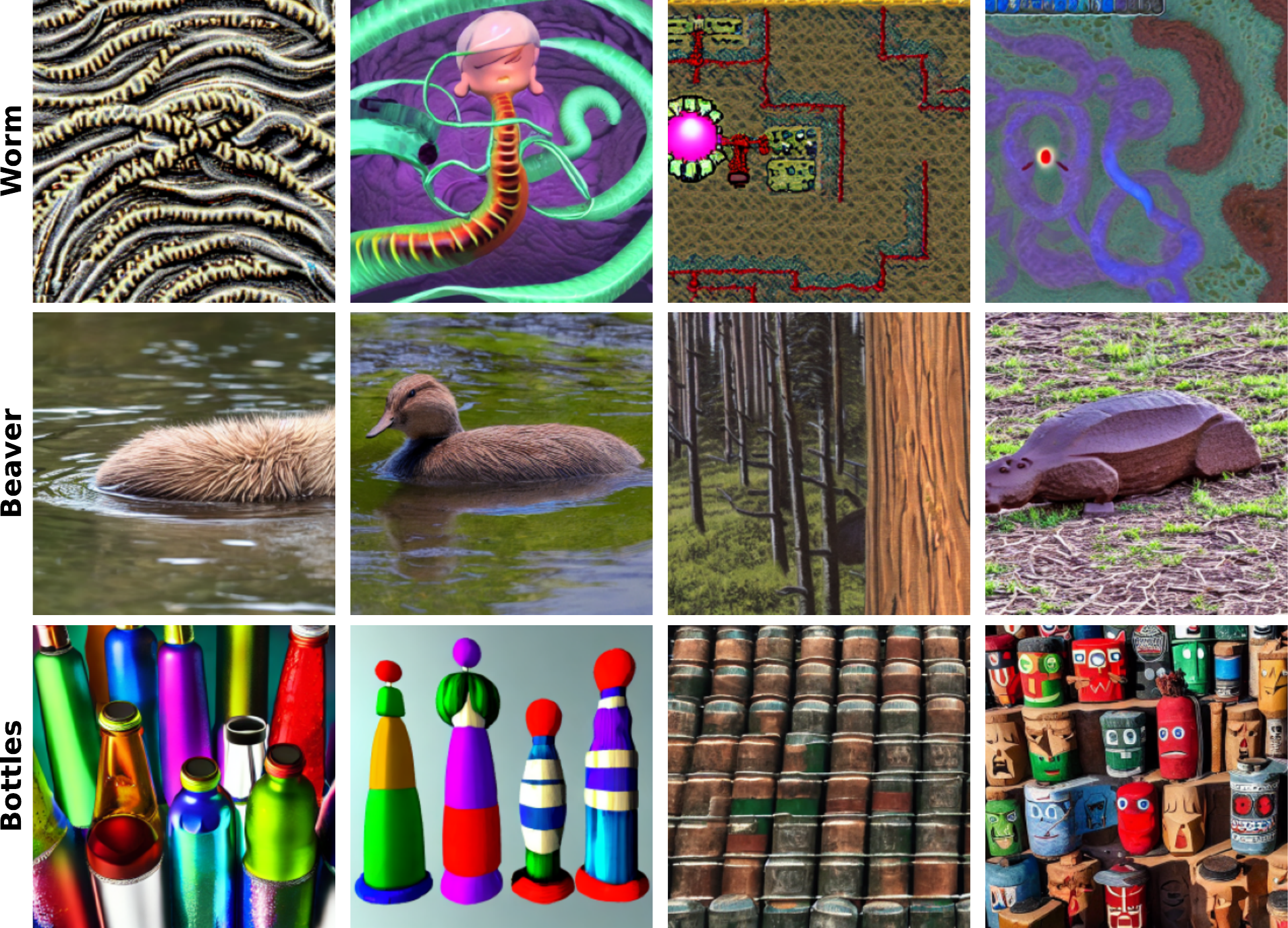}
    \caption{\textbf{Random outliers generated for three CIFAR100 classes by our method.} Using our generated samples as auxiliary outliers during training greatly improves the OOD detection performance of modern classifiers.}
    \label{fig:teaser}
\end{figure}

To identify samples that do not belong to the training distribution, OOD detectors seek to learn a boundary that separates in-distribution (ID) samples from OOD samples. A major difficulty in doing so is the lack of real OOD samples at training time. Recent advancements in supervised OOD detection tackle this challenge by generating synthetic OOD samples and using them during training to shape the decision boundary. These synthetic OOD samples can be generated in feature space~\cite{du2022towards,tao2023nonparametric} or pixel space~\cite{chen2024fodfom,du2024dream}, with pixel-space approaches demonstrating superior performance. However, creating realistic and effective pixel-space OOD samples remains challenging. The standard approach~\cite{chen2024fodfom,du2024dream} is to use large, pre-trained text-conditioned diffusion models, like Stable Diffusion, conditioned on perturbed prompts that lead to OOD images for specific classes.

\begin{figure*}[t]
  \centering
  \setlength\tabcolsep{2pt}
  \begin{tabular}{ccc}
    \includegraphics[width=0.3\linewidth]{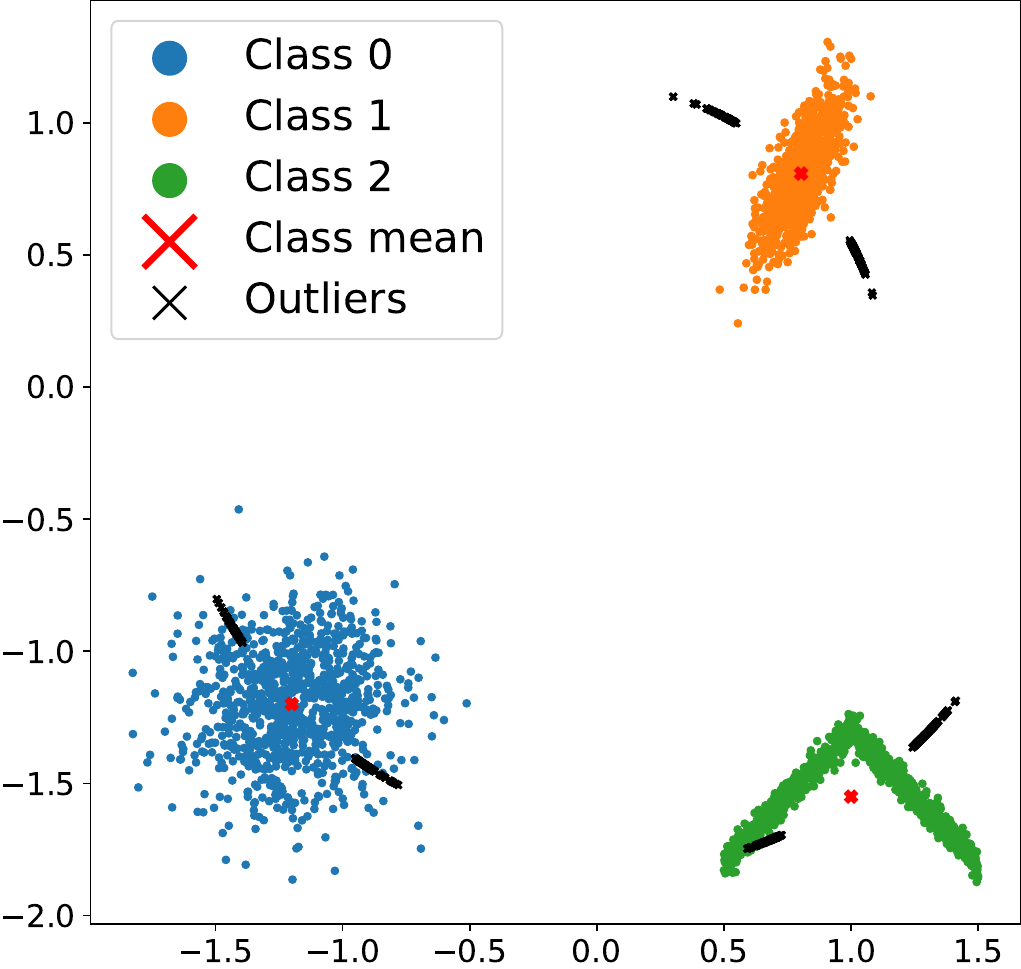} &
    \includegraphics[width=0.3\linewidth]{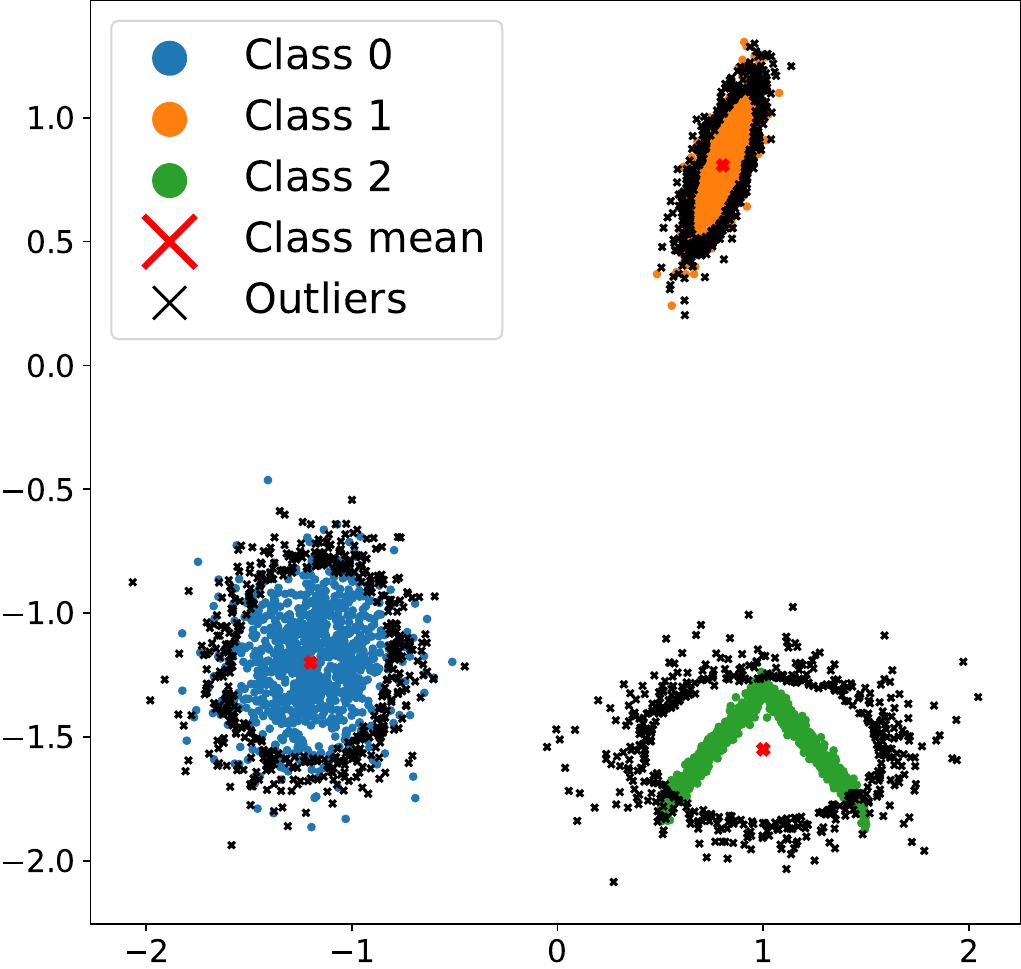} &
    \includegraphics[width=0.3\linewidth]{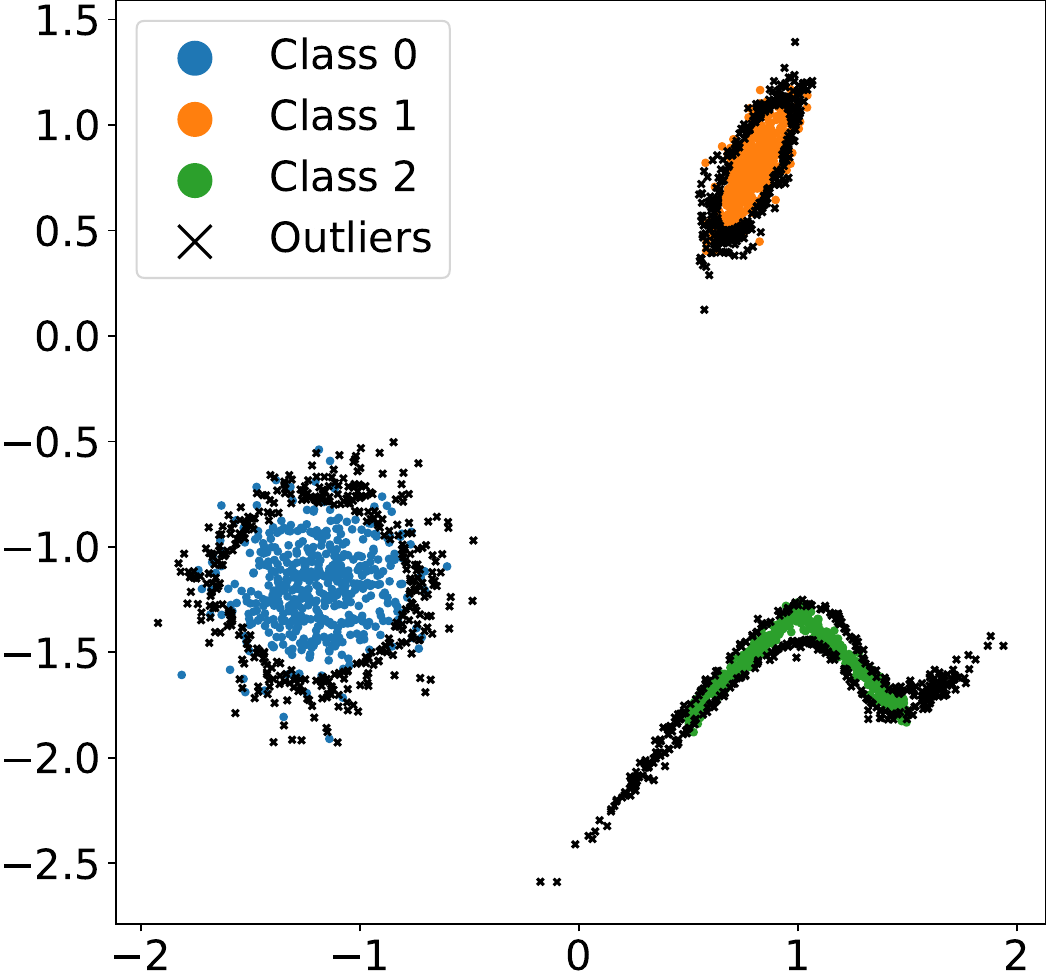} \\
    (a) & (b) & (c) \\
  \end{tabular}
    \caption{\textbf{Comparison between (a) Dream-OOD, (b) linear invariants on our embeddings, and (c) our proposed method on a toy example.} The disjoint embeddings and normalization of Dream-OOD greatly limit the flexibility of the generated outliers. Linear invariants similarly lack capacity. On the other hand, our method can generate successful outliers by modeling arbitrary distributions. }
    \label{fig:toy}
\end{figure*}

Different types of prompt perturbations produce outlier images of distinct quality and, therefore, finding good perturbations is crucial to generate useful OOD samples. Ideally, the perturbations should locate the prompts in the boundaries of the diffusion model's conditioning space where the model transitions from producing valid ID images to OOD images (\eg,~Fig.~\ref{fig:toy}(c)).
However, previous pixel-space outlier synthesis methods have overlooked the key challenge of identifying such boundaries within the diffusion model's conditioning space, instead relying on heuristics and approximations that fall short in effectively locating these regions.
For example, Dream-OOD~\cite{du2024dream}, the current state-of-the-art method, trains an image encoder that maps ID images to a feature space encouraging that each image embedding is similar to the text embedding of its respective class name. Yet, this approach lacks a mechanism to ensure alignment between the learned image embeddings and the diffusion model's conditioning space. Consequently, embeddings considered as ID by the embedding model may lie in OOD regions of the conditioning space, and vice versa. Moreover, Dream-OOD fits a class-conditional spherical Gaussian distribution to the ID image embeddings, but there is no evidence that the prompts embeddings in the conditioning space behave in this manner.

In this work, we develop two new techniques to explicitly find ID regions within the diffusion model's conditioning space and to sample elements along their boundaries:
\begin{itemize}
    \item \textbf{Diffusion-based embedding.} We propose an embedding procedure that uses the diffusion model directly to create image embeddings, bypassing the need for external embedding functions or surrogate models. This approach, similar to prompt tuning, represents each training image by an embedding that maximizes the likelihood of the diffusion model generating that image, allowing a precise characterization of the ID regions within the conditioning space.
    \item \textbf{Non-linear parametric distributions.} Inspired by the non-linear invariants for OOD detection of~\cite{doorenbos2024learning}, we introduce a conditional volume-preserving network~(cVPN) for fitting class-wise manifolds and show how it can be used to fit arbitrarily complex class-conditional distributions to the training image embeddings. The complex distributions enabled by the cVPN allow for more precise sampling along the ID/OOD boundaries in the conditioning space compared to the over-simplistic Gaussian distribution (Fig.~\ref{fig:toy}(a)~and~(b)).
\end{itemize}
By combining these two techniques, our proposed method \framework{} can synthesize realistic and diverse outliers, as illustrated in Fig.~\ref{fig:teaser}, thereby enhancing the classifier's capacity to detect OOD samples. Our experiments show that \framework{} achieves state-of-the-art performance on two canonical benchmarks (ImageNet-100 and CIFAR-100 as in-distributions), with ablation studies confirming the individual impact of each component. Additionally, we highlight a limitation of outlier synthesis methods, revealing a bias toward low-level image statistics. We demonstrate that this can be mitigated effectively through tailored data augmentation.

\section{Related Work}
Many works on supervised OOD detection, including most early ones, base their scoring function on the post-hoc processing of the classifier. These can, for instance, be based directly on the logits~\cite{hendrycks2016baseline,hendrycks2019scaling} or processed versions thereof~\cite{liang2018enhancing,liu2020energy,liu2023gen}, the features learned by the model~\cite{kamoi2020mahalanobis,lee2018simple,ren2021simple,sastry2020detecting,sun2022out}, or the gradients of the classifier~\cite{behpour2024gradorth,huang2021importance}. Other approaches adapt the classifier training process with OOD detection in mind. This can be done by modifying the loss function, such as by learning to estimate the confidence in a given prediction and rejecting samples when the confidence is low~\cite{devries2018learning,hsu2020generalized,lee2017training}, training with an energy-based loss~\cite{liu2020energy}, or by adding auxiliary self-supervised objectives~\cite{ahmed2020detecting,winkens2020contrastive}. 

More recently, a popular trend is to use large auxiliary datasets as fake OOD samples during training, a technique known as Outlier Exposure (OE)~\cite{hendrycks2018deep}. The rationale behind this approach is that by exposing the model to a diverse set of general samples — such as from ImageNet when dealing with natural images — it can better learn to recognize what it does not know. 
While OE has been successfully applied to increase the performance of many OOD detectors (e.g.,~\cite{fort2021exploring,hendrycks2018deep,papadopoulos2021outlier,sastry2020detecting}), its usefulness is limited when the OE distribution is far from the OOD distribution~\cite{reiss2023mean}, and, for many domains, obtaining a relevant large and diverse dataset to use as pseudo-OOD samples is infeasible. 

As such, current approaches rely only on in-distribution data to build their detectors. The state-of-the-art approaches opt to synthesize outliers, which are used to regularize the classification model during training to improve its OOD detection. VOS~\cite{du2022towards} and NPOS~\cite{tao2023nonparametric} do so in the feature space, while more recent methods~\cite{chen2024fodfom,du2024dream} generate outliers in pixel space and show its superior performance. Our approach also generates outliers in pixel space, allowing for interpretability, but we build upon previous work by aligning the embeddings and modeling non-linearities, allowing for a significant performance improvement. Similar ideas can also be found in anomaly segmentation, where OOD patches are pasted into images and should be correctly identified~\cite{li2021cutpaste,schluter2022natural,tan2021detecting}.

In contrast, the unsupervised OOD detection literature, where no class labels are available, has many different methods typically based on generative models~\cite{nalisnick2018deep,schirrmeister2020understanding,serra2019input}, self-supervised learning~\cite{abati2019latent,hendrycks2019using,sehwag2021ssd}, or pre-trained models~\cite{doorenbos2022data,reiss2023mean,xiao2021we}. However, their cross-domain compatibility is rarely explored. Relevant to the present work is the concept of data invariants~\cite{doorenbos2022data}, which characterizes what makes a sample in-distribution without labels, and follow-up work on learning non-linear invariants~\cite{doorenbos2024learning}, where a network was developed that can learn non-linear relations that collectively describe a training dataset. We are the first to bring this concept to supervised OOD detection and introduce appropriate modifications to adapt these methods to this new context.

Finally, we make heavy use of diffusion models in our work. Diffusion models are generative models capable of generating high-quality samples resembling their training dataset. These models have found success in a large number of applications, including image generation~\cite{dhariwal2021diffusion}, dataset building~\cite{wu2023datasetdm}, industrial anomaly detection~\cite{fuvcka2025transfusion}, sampling from low-density regions~\cite{sehwag2021ssd,um2024self}, and more. Many of these advancements were made possible by large pre-trained diffusion models, such as Stable Diffusion~\cite{rombach2021highresolution}, which we also rely on in our work. In particular, we build a training set representation in diffusion embedding space using techniques similar to those in personalized text-to-image generation (e.g.,~\cite{gal2022image}). However, we use this to represent the manifold of a dataset rather than single concepts.
\section{Method}

OOD detection aims to determine whether a given test sample originates from the same distribution as the training data. Samples drawn from the training distribution are considered in-distribution~(ID), while those that deviate are considered out-of-distribution~(OOD). OOD detection can be formulated as finding a scoring function~$s: \mathcal{X}\to\real$ that assigns a score of \emph{in-distributionness}~$s(\x)$ to each input~$\x$. In the context of supervised OOD detection, where the downstream task involves a classifier trained on a labeled dataset, the OOD detection is typically integrated into the classifier by introducing a training regularization term that encourages higher free energy to ID samples and lower free energy to OOD samples, effectively using the free energy as the OOD score function. However, to achieve this energy separation, the regularization term requires ID samples, available in the training data, as well as OOD samples, which must be artificially generated. Our method aims to produce high-quality OOD images that, when used by the regularization term at training time, boost the robustness of the classifier to OOD samples. The following sections describe the components of our method and how they are combined.

\begin{algorithm}[t!]
\caption{Obtaining token embedding for an image $\x$.}
\label{alg:embed}
\begin{algorithmic}
\Require Training image $\x$, label embedding $\e_\ell$, number of iterations, batch size, learning rate $\eta$, trained diffusion model $\epsilon_\theta$.
\State $\e \gets \e_\ell$ \Comment{Initialize with the token embedding for label}
\For{number of iterations}
    \LineComment{Sample $t$ and $\epsilon$ for the entire batch}
    \State $t \sim \textrm{Uniform}(\{1,...,T\}, \text{batch size})$ 
    \State $\epsilon \sim \mathcal{N}(\bm{0}, \bm{I}, \text{batch size})$ 
    \LineComment{Loss calculation with Eq.~\eqref{eq:ddpm_loss}}
    \State $L = \| \epsilon - \epsilon_\theta(\hat{\x}_t, t, \e)\|^2$ 
    \State $\e \gets \e - \eta \nabla_{\e} L$ \Comment{Gradient descent}
\EndFor
\State \Return $e$
\end{algorithmic}
\end{algorithm}

\begin{figure*}
    \centering
    \includegraphics[width=0.9\linewidth]{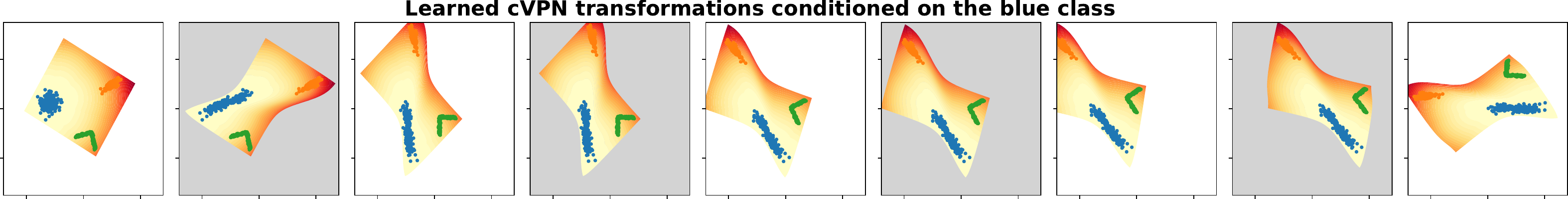} \\
    \includegraphics[width=0.9\linewidth]{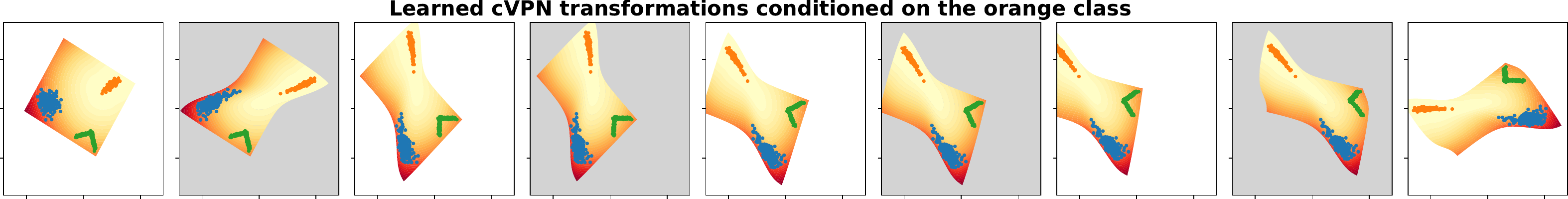} \\
    \includegraphics[width=0.9\linewidth]{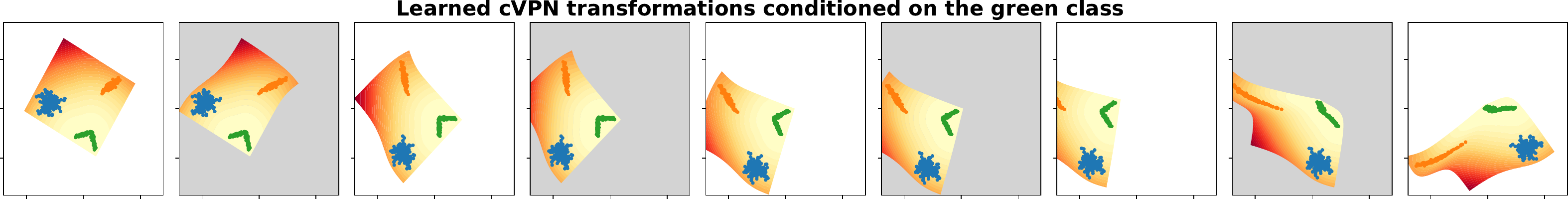} \\
    \caption{\textbf{The class-specific representations learned by the cVPN on toy data with three classes.} Depending on the conditioning, the cVPN transforms the input data such that the current class is transformed into a representation with an invariant (the y-axis). The background color indicates the distance to the nearest training data point from the current class in the original space. Images with a white-shaded background result from rotation layers, and images with a gray background result from the conditional coupling layers.}
    \label{fig:enter-label}
\end{figure*}

\subsection{Embedding images in the diffusion conditioning space}
\label{sec:embedding}

We embed the training images directly within the native conditioning space of Stable Diffusion, which we call the \emph{diffusion conditioning space} or simply \emph{diffusion space}. For each training image-label pair~$(\x, \ell)$, we derive an embedding~$\e$ by minimizing the standard noise-prediction loss used in diffusion models with respect to the condition vector~$\e$,
\begin{equation}
\label{eq:ddpm_loss}
    \argmin_{\e} \EX_{t,\epsilon\sim\mathcal{N}(\mathbf{0}, \mathbf{I})} \left\| \epsilon - \epsilon_\theta(\hat{\x}_t, t, \e)\right\|^2 + R(\e, \mathbf{e}_\ell),
\end{equation}
where $\hat{\x}_t$ is the noisy version of~$\x$ with noise~$\epsilon$ and $\mathbf{e}_\ell$~is the Stable Diffusion token embedding for the label~$\ell$. The regularization term~$R$ encourages~$\e$ to remain close to the token embedding for the label~$\ell$. In practice, we implement this regularization term by initializing $\e$ to $\mathbf{e}_\ell$ and minimizing only the first term in Eq.~\eqref{eq:ddpm_loss} over a limited number of iterations (set to~3 in our experiments). The process is summarized in Alg.~\ref{alg:embed}. We apply this embedding process across the training set to obtain a collection of \emph{diffusion embeddings}~$\{\e_i\}_{i=1}^N$.

Formally, the diffusion embedding~$\e$ of an image~$\x$ is a maximum-a-posteriori estimate that maximizes the likelihood of Stable Diffusion generating the same image~$\x$, with the regularization term acting as the prior. Therefore, the set of training diffusion embeddings~$\{\e_i\}_{i=1}^N$ forms a non-parametric distribution capturing the regions in the diffusion conditioning space where Stable Diffusion is most likely to produce in-distribution (ID) samples. The optimal OOD samples lie in the boundaries of these regions. In the following sections, we describe how to represent these regions using non-linear parametric distributions for easy sampling along the region boundaries.

\subsection{Fitting class-wise non-linear manifolds to ID data}
\label{sec:nonlinear}

As introduced in~\cite{doorenbos2022data,doorenbos2024learning}, the manifolds where the in-distribution (ID) data lies can be effectively modeled by identifying functions, or \emph{invariants}, that remain approximately constant across the ID samples. These invariants capture essential properties of the ID data, remaining stable for ID samples but diverging for OOD samples. In particular, the non-linear invariants~(\NL) introduced in~\cite{doorenbos2024learning} find ID data invariants~$\g$ within the latent space by solving
\begin{align}
    \label{eq:soft_invariant_g}
    \min_\g & \sum_i \|\g(\e_i)\|^2_2 \\
    \label{eq:soft_independent_g}
    \textrm{s.t.} & \det(\J(\e_i)\cdot\J^T(\e_i)) \neq 0 \quad \forall i,
\end{align}
where the constraint ensures that the Jacobian is full-rank. This full-rank condition is achieved by design through a \emph{volume preserving network}~(VPN), constructed as a sequence of bijective layers (interleaved orthogonal and coupling layers) with unimodular Jacobians. During training, the VPN minimizes only the primary term in Eq.~\eqref{eq:soft_invariant_g}, and its volume-preserving structure prevents the network from collapsing to a trivial (near-)constant projection and artificially minimizing Eq.~\eqref{eq:soft_invariant_g}. 

We extend the original formulation of \NL{} to a supervised OOD detection framework, allowing the model to learn class-conditional manifolds within the space of the diffusion embeddings. To do so, we introduce a conditional volume-preserving network (cVPN), which implements a function~$\f: \real^D\times\mathcal{Y} \mapsto \real^D$ that is bijective with respect to its first argument and conditioned on the second argument representing the class. The first $K<D$ output dimensions of~$\f$ correspond to the invariants, $\g=\f_{1:K}$, while the remaining dimensions model the variability within the ID data. Fitting the cVPN is done by optimizing a problem analog to Eqs.~\eqref{eq:soft_invariant_g} and~\eqref{eq:soft_independent_g},
\begin{align}
    \label{eq:cond_soft_invariant_g}
    \min_\g & \sum_i \|\g(\e_i, \ell_i)\|^2_2 \\
    \label{eq:cond_soft_independent_g}
    \textrm{s.t.} & \det(\J(\e_i, \ell_i)\cdot\J^T(\e_i, \ell_i)) \neq 0 \quad \forall i.
\end{align}
The trained cVPN thus maps elements from the diffusion space to an \emph{invariant space}, ensuring that $\f_k(\e_i, \ell) \approx \f_k(\e_j, \ell)\approx 0$ for all $k<K$ when $\e_i$ and $\e_j$ are diffusion embeddings of images from the same class~$\ell$. 

To integrate class-specific information effectively, the cVPN replaces the original coupling layers from~\cite{doorenbos2024learning} with \emph{conditional coupling layers} that incorporate class information through a learnable embedding function~$h_\theta: \mathcal{Y} \mapsto \real^{\ceil{D/2}}$. This function maps the class label~$\ell$ into a feature space of dimension~$\ceil{D/2}$, enabling the conditional coupling layers to adapt based on class context. The class embedding is passed to the MLP~$t$ of the conditional coupling layer. The conditional coupling layer is thus defined as
\begin{align}
    \textrm{ccl}(\x, \ell) &= \textrm{join}\left(\x_{1:d} + t(\x_{d+1:D}, h_\theta(\ell)), \x_{d+1:D}\right),
\end{align}
where~$d$ is typically set to $\ceil{D/2}$. Its inverse is
\begin{align}
    \textrm{ccl}^{-1}(\x, \ell) &= \textrm{join}\left(\x_{1:d} - t(\x_{d+1:D}, h_\theta(\ell)), \x_{d+1:D}\right).
\end{align}
We show how a cVPN works on a toy example in Fig.~\ref{fig:enter-label}.

\subsection{Probability distribution of ID samples}

\begin{figure}
    \centering
    \includegraphics[width=0.9\linewidth]{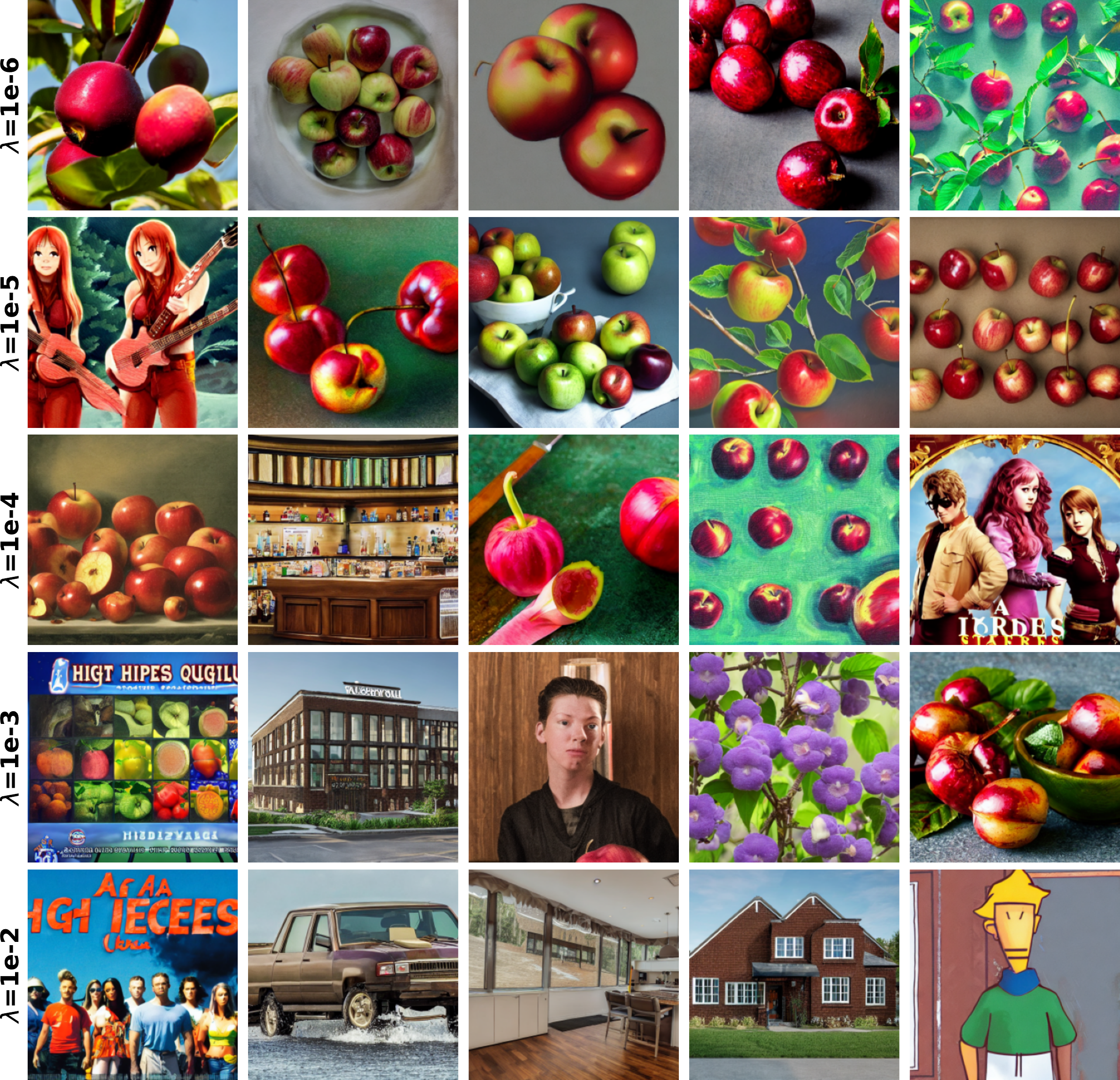}
    \caption{\textbf{Effect of regularization on generated outliers.} Outliers get progressively more OOD with stronger regularization, providing an intuitive way to control their difficulty.}
    \label{fig:reg}
\end{figure}

We leverage the bijective nature of the cVPN network to fit non-linear parametric distributions to the ID diffusion embeddings. Specifically, we map the diffusion embeddings to the invariant space evaluating $\bv_i = \f(\e_i, \ell_i)$ for each embedding~$\e_i$ and its corresponding label~$\ell_i$. This process yields a collection of \emph{invariant-space vectors}~$\{\bv_i\}_{i=1}^N$. We then fit class-conditional Gaussian distributions to these invariant-space vectors for each class~$\ell$,
\begin{align}
    p_v(\bv_i \mid \ell) &\sim \mathcal{N}(\bv_i; \bm{\mu}_\ell, \bm{\Sigma}_\ell + \lambda \textbf{I}), \\
    \bm{\mu}_\ell &= \frac{1}{N_\ell} \sum_{i:\ell_i=\ell} \bv_i,\\
    \bm{\Sigma}_\ell &= \frac{1}{N_\ell} \sum_{i:\ell_i=\ell} (\bv_i - \bm{\mu}_\ell)(\bv_i - \bm{\mu}_\ell)^\top,
\end{align}
where the regularization factor~$\lambda$ applied to the covariance matrix serves two primary purposes. First, it prevents numerical issues that might arise from the invariant dimensions approaching near-zero values. Second, it enables control over the degree of \textit{out-of-distributionness} of the generated outliers. By setting $\lambda$ to be small relative to the variant dimensions but large relative to the invariant dimensions, higher values of $\lambda$ yield outliers with more extreme values in the invariant dimensions, as illustrated in Fig.~\ref{fig:reg}. We ablate the effect of~$\lambda$ in Sec.~\ref{sec:disc}.

The class-conditional Gaussian distributions defined in the invariant space induce non-linear class-conditional probability density functions in the diffusion space,
\begin{equation}
    p_e(\e\mid \ell) = p_v(f(\e, \ell)\mid \ell) \cdot \left| \J(\e, \ell) \right|,
\end{equation}
where the determinant of the Jacobian is 1 by design, resulting in the simplified expression $p_e(\e\mid \ell) = p_v(f(\e, \ell)\mid \ell)$.

\subsection{OOD sample generation}

Generating an OOD image for a given class~$\ell$ follows naturally from the properties of the distributions established above. First, we apply rejection sampling in the invariant space to obtain an outlier~$\bv'$ from the low-likelihood regions of the Gaussian distribution $p_v(\bv'\mid \ell)$. Then, we map this outlier back to the diffusion space using the inverse of the cVPN~$\f$, yielding an embedding $\e'=\f^{-1}(\bv'; \ell)$. Finally, we condition the Stable Diffusion model on prompts of the form ``A high-quality image of a $\langle \e' \rangle$'' to generate an OOD image~$\x'$. The entire process is denoted as $\x'\sim \mathbb{P}_\textrm{out}$.

\begin{table*}[t]
  \centering
  \small
  \resizebox{\textwidth}{!}{
    \begin{tabular}{cccccccccccccc}
    \toprule
     \multirow{2}[4]{*}{Methods} & \multicolumn{2}{c}{\textsc{Svhn}} & \multicolumn{2}{c}{\textsc{Places365}} & \multicolumn{2}{c}{\textsc{Lsun}} & \multicolumn{2}{c}{\textsc{iSun}} & \multicolumn{2}{c}{\textsc{Textures}} & \multicolumn{2}{c}{Average} & \multirow{2}[4]{*}{Acc} \\
\cmidrule{2-13}          &FPR95$\downarrow$ & AUC$\uparrow$ & FPR95$\downarrow$ & AUC$\uparrow$ & FPR95$\downarrow$ & AUC$\uparrow$ & FPR95$\downarrow$ & AUC$\uparrow$ & FPR95$\downarrow$ & AUC$\uparrow$  & FPR95$\downarrow$ & AUC$\uparrow$   \\
    \midrule
    MSP~\cite{hendrycks2016baseline}   &  87.35 & 69.08 & 81.65 & 76.71  & 76.40 & 80.12& 76.00 & 78.90 & 79.35 & 77.43 & 80.15 & 76.45  & 79.04 \\
    ODIN~\cite{liang2018enhancing}  & 90.95 & 64.36 & 79.30 & 74.87 & 75.60 & 78.04& 53.10 & 87.40  & 72.60 & 79.82 & 74.31 & 76.90  & 79.04\\
    Mahalanobis~\cite{lee2018simple}& 87.80 & 69.98 & 76.00 & 77.90  & 56.80 & 85.83 & 59.20 & 86.46& 62.45 & 84.43& 68.45 & 80.92  & 79.04\\
    Energy~\cite{liu2020energy} & 84.90 & 70.90 & 82.05 & 76.00& 81.75 & 78.36 & 73.55 & 81.20 &  78.70 & 78.87  & 80.19 & 77.07 & 79.04\\
    G-ODIN~\cite{hsu2020generalized} & 63.95 & 88.98 & 80.65 & 77.19  & 60.65 & 88.36  &  51.60 & 92.07 & 71.75 & 85.02  & 65.72 & 86.32 & 76.34\\
   kNN~\cite{sun2022out}   &81.12 & 73.65   & 79.62 & 78.21 & 63.29 & 85.56 & 73.92 & 79.77& 73.29 & 80.35 & 74.25 & 79.51 & 79.04 \\
  ViM~\cite{wang2022vim}   &   81.20 & 77.24 &  79.20 & 77.81  & 43.10 & 90.43& 74.55 & 83.02 & 61.85 & 85.57 &  67.98 & 82.81  & 79.04 \\
  ReAct~\cite{sun2021react} & 82.85 & 70.12 & 81.75 & 76.25 & 80.70 & 83.03 &67.40 & 83.28 &74.60 & 81.61 & 77.46 & 78.86 &79.04\\
DICE~\cite{sun2022dice}& 83.55 & 72.49  & 85.05 & 75.92 & 94.05 & 73.59 &75.20 & 80.90 &79.80 & 77.83 &83.53 & 76.15 & 79.04 \\
\hline
\emph{Synthesis methods}\\
 VOS~\cite{du2022towards}   & 78.50 & 73.11& 84.55 & 75.85 & 59.05 & 85.72  & 72.45 & 82.66 & 75.35 & 80.08&  73.98 & 79.48& 78.56\\
 NPOS~\cite{tao2023nonparametric}  & \textbf{11.14} & \textbf{97.84} & 79.08 & 71.30 &  56.27 & 82.43 & 51.72 & 85.48 & \underline{35.20} &  \underline{92.44} & 46.68 &   85.90 & 78.23\\
 Dream-OOD~\cite{du2024dream}  & 58.75 & 87.01 & \underline{70.85} & \underline{79.94} & \underline{24.25} & \underline{95.23} & \textbf{1.10} & \textbf{99.73} & 46.60 & 88.82 &40.31 & 90.15 & 78.94\\
 \rowcolor{mygray} \textbf{\framework} (ours) & \underline{14.43}\tiny{$\pm3.5$} & 	\underline{96.76}\tiny{$\pm0.8$} & 	\textbf{8.72}\tiny{$\pm0.5$} & 	\textbf{97.71}\tiny{$\pm0.2$} & 	\textbf{21.72}\tiny{$\pm3.1$} & 	\textbf{95.39}\tiny{$\pm0.5$} & 	\underline{1.42}\tiny{$\pm0.5$} & 	\underline{99.56}\tiny{$\pm0.1$} & 	\textbf{7.9}\tiny{$\pm0.5$} & 	\textbf{97.96}\tiny{$\pm0.3$} & 	\textbf{10.84}\tiny{$\pm0.8$} & 	\textbf{97.48}\tiny{$\pm0.2$} & 	78.86\tiny{$\pm0.5$} \\
   \hline
\end{tabular}}
 \caption{\textbf{Comparative evaluation with \textsc{Cifar-100} as the in-distribution data.} \textbf{Bold} and \underline{underlined} indicate best and second best per column, respectively. Our results are over 3 seeds. Baseline performances taken from~\cite{du2024dream}.}
    \label{tab:ood_results_c100}
\end{table*}

\subsection{Classifier regularization with synthetic OOD samples}
\label{sec:regularization}

Synthetic OOD samples, together with real ID samples from the training data, are used to regularize the classifier, enhancing its ability to distinguish ID from OOD inputs as proposed in~\cite{du2022towards,du2024dream,tao2023nonparametric}. This approach combines the standard cross-entropy loss $\mathcal{L}_\text{CE}$ with an additional regularization term, $\mathcal{L}_\text{ood}$, yielding
\begin{equation}
    \mathcal{L} = \mathcal{L}_\text{CE}+\beta \cdot \mathcal{L}_\text{ood},
\end{equation}
where the hyperparameter~$\beta$ controls the influence of the regularization. The regularization term~$\mathcal{L}_\text{ood}$ is designed to encourage distinct classifier energy levels for ID and OOD samples,
\begin{align}
    \nonumber
\mathcal{L}_\text{ood}=\mathbb{E}_{\x \sim \mathbb{P}_{\text{out}}}&\left[-\log \frac{1}{1+\exp(\phi \circ E \circ f_{\theta}(\x))}\right]\\
    + \mathbb{E}_{\x \sim \mathbb{P}_\text{in}}&\left[-\log \frac{\exp(\phi \circ E \circ f_{\theta}(\x))}{1+\exp(\phi \circ E \circ f_{\theta}(\x))}\right].
\end{align}
Here, the classifier $f_\theta$ processes the input image~$\x$ to produce the class logits for labels in~$\mathcal{Y}$. The energy function~$E$, as defined in~\cite{liu2020energy}, transforms these logits into an energy score that reflects the model's certainty about a given sample being ID or OOD. The function~$\phi$, implemented as an MLP, transforms energy values into logits, effectively classifying each sample~$\x$ as either \texttt{ID} or \texttt{OOD}. Thus, the composition $s = \phi \circ E \circ f_\theta$ serves as the OOD scoring function at inference time. We refer to our full framework as Non-Linear Class-wise Invariant Sampling, or \framework for short.
\section{Experiments}

Following previous works~\cite{du2022towards,du2024dream}, we conduct experiments on two benchmarks:
\begin{description}
    \item[CIFAR-100~\cite{krizhevsky2009learning}] as ID dataset with SVHN~\cite{netzer2011reading}, Places365~\cite{zhou2017places}, LSUN~\cite{yu2015lsun}, iSun~\cite{xu2015isun}, and Textures~\cite{cimpoi2014describing} as OOD datasets;
    \item[ImageNet-100~\cite{deng2009imagenet}] as the ID dataset with iNaturalist~\cite{van2018inaturalist}, Places~\cite{zhou2017places}, Sun~\cite{xiao2010sun}, and Textures~\cite{cimpoi2014describing} as OOD.
\end{description}
See the appendix for more details. We report the false positive rate at 95\% true positive rate~(FPR95) and the area under the receiver operating characteristic curve~(AUC) for the task of OOD detection, with the positive class being~\texttt{ID}. We also report the accuracy~(Acc) of the classifier on the downstream classification task.

We follow the training protocol described in~\cite{du2024dream} to isolate the effects of our outlier synthesis approach from other factors on the OOD detection task. We train a ResNet-34 with stochastic gradient descent, using a learning rate of $10^{-1}$ for CIFAR100 and $10^{-3}$ for ImageNet100, decaying with a cosine annealing schedule, momentum of 0.9, weight decay of $5\cdot{}10^{-4}$, and a batch size of~160. We also use Stable Diffusion~v1.4, using DDIM sampling with 50~steps, to generate the synthetic outliers and set~$\beta$~to~1.0. 

The classifier is trained for 20~epochs on ImageNet-100 and 250~epochs on~CIFAR100. Training embeddings are obtained by optimizing Eq.~\eqref{eq:ddpm_loss} for three iterations with a batch size of~32, ensuring sufficient exposure to different DDIM steps. We set the number of invariants~$K$ to the average largest number of principal components that jointly account for less than $p\%$~of the variance per class~\cite{doorenbos2024learning}. We set $p=2$ and a regularization strength of~$\lambda=10^{-5}$ in our experiments, and then we analyze the impact of different values of these hyperparameters in the ablation study.

Following~\cite{du2024dream}, we compare \framework against a suite of strong baselines: MSP~\cite{hendrycks2016baseline}, ODIN~\cite{liang2018enhancing}, Mahalanobis~\cite{lee2018simple}, Energy~\cite{liu2020energy}, G-ODIN~\cite{hsu2020generalized}, kNN~\cite{sun2022out}, ViM~\cite{wang2022vim}, ReAct~\cite{sun2021react}, DICE~\cite{sun2022dice}, VOS~\cite{du2022towards}, NPOS~\cite{tao2023nonparametric}, and Dream-OOD~\cite{du2024dream}. All methods use only in-distribution data with auxiliary outlier datasets for a fair comparison. As we use the same code-base and training settings from~\cite{du2024dream}, we report baseline numbers from their experiments.

\begin{table*}[t]
  \centering
 \resizebox{\textwidth}{!}{
    \begin{tabular}{cccccccccccc}
    \toprule
          \multirow{2}[4]{*}{Methods} & \multicolumn{2}{c}{\textsc{iNaturalist}} & \multicolumn{2}{c}{\textsc{Places}} & \multicolumn{2}{c}{\textsc{Sun}} & \multicolumn{2}{c}{\textsc{Textures}}  & \multicolumn{2}{c}{Average} & \multirow{2}[4]{*}{Acc} \\
\cmidrule{2-11}          & FPR95$\downarrow$ & AUC$\uparrow$ & FPR95$\downarrow$ & AUC$\uparrow$ & FPR95$\downarrow$ & AUC$\uparrow$ & FPR95$\downarrow$ & AUC$\uparrow$ & FPR95$\downarrow$ & AUC$\uparrow$ &  \\
\hline

MSP~\cite{hendrycks2016baseline}& 31.80 & 94.98 & 47.10 & 90.84  &47.60 & 90.86 & 65.80 & 83.34 &  48.08 & 90.01& 87.64  \\
ODIN~\cite{liang2018enhancing}& 24.40 & 95.92& 50.30 & 90.20& 44.90 & 91.55 &61.00 & 81.37&  45.15 & 89.76 & 87.64\\
Mahalanobis~\cite{lee2018simple}& 91.60 & 75.16& 96.70 & 60.87&  97.40 & 62.23& 36.50 & 91.43& 80.55 & 72.42& 87.64 \\
Energy~\cite{liu2020energy}& 32.50 & 94.82& 50.80 & 90.76 & 47.60 & 91.71  & 63.80 & 80.54 & 48.68 & 89.46 & 87.64\\
G-ODIN~\cite{hsu2020generalized} & 39.90 & 93.94 & 59.70 & 89.20 &58.70 & 90.65 & 39.90 & 92.71 & 49.55 & 91.62 & 87.38\\
kNN~\cite{sun2022out}& 28.67 & 95.57& 65.83 & 88.72& 58.08 & 90.17& {12.92} & 90.37&  41.38 & 91.20& 87.64  \\
ViM~\cite{wang2022vim}& 75.50 & 87.18& 88.30 & 81.25& 88.70 & 81.37& 15.60 & {96.63}& 67.03 & 86.61& 87.64 \\
ReAct~\cite{sun2021react} &22.40 & 96.05 &  45.10 & 92.28  & 37.90 & 93.04 & 59.30 & 85.19  & 41.17 & 91.64  & 87.64 \\
DICE~\cite{sun2022dice}& 37.30 & 92.51 & 53.80 & 87.75 &45.60 & 89.21 & 50.00 & 83.27 &46.67 & 88.19 &87.64 \\
\midrule
\emph{Synthesis methods}\\
VOS~\cite{du2022towards}&  43.00 & 93.77 &47.60 & 91.77 &39.40 & 93.17 &66.10 & 81.42 &  49.02 & 90.03 &87.50\\
 
 NPOS~\cite{tao2023nonparametric}  &53.84 & 86.52 &  59.66 & 83.50 & 53.54 & 87.99 & \textbf{8.98} &  \textbf{98.13} & 44.00 &  89.04 &  85.37\\
  Dream-OOD~\cite{du2024dream} & \underline{24.10} & \underline{96.10}  & \underline{39.87} & \underline{93.11} & \underline{36.88} & \underline{93.31}  &  53.99 & 85.56  & \underline{38.76} & \underline{92.02} & 87.54 \\
\rowcolor{mygray} \textbf{\framework} (ours) & \textbf{20.7}\tiny{$\pm0.2$} & 	\textbf{96.56}\tiny{$\pm0.2$} & 	\textbf{34.6}\tiny{$\pm0.4$} & 	\textbf{94.07}\tiny{$\pm0.2$} & 	\textbf{35.43}\tiny{$\pm0.8$} & 	\textbf{94.13}\tiny{$\pm0.2$} & 	\underline{44.83}\tiny{$\pm1.8$} & 	\underline{88.5}\tiny{$\pm0.9$} & 	\textbf{33.89}\tiny{$\pm0.6$} & 	\textbf{93.32}\tiny{$\pm0.2$} & 	87.24\tiny{$\pm0.1$}  \\
            \hline
    \end{tabular}%
    }
     \caption{\textbf{Comparative evaluation with \textsc{ImageNet-100} as the in-distribution data.} \textbf{Bold} and \underline{underlined} indicate best and second best per column, respectively. Our results are over 3 seeds. Baseline performances taken from~\cite{du2024dream}.}
  \label{tab:ood_results_in100}%
\end{table*}

\section{Discussion}
\label{sec:disc}

We find that \framework method reaches a new state-of-the-art (Tab.~\ref{tab:ood_results_c100} and Tab.~\ref{tab:ood_results_in100}), surpassing the best FPR95 by \textbf{29.47} and~\textbf{4.87} on CIFAR100 and ImageNet100, respectively. \framework achieves the best performance on six out of the nine experiments, placing second in the remaining three, thereby outperforming all traditional, non-synthesis-based methods. 

We also improve upon all other outlier synthesis methods, whether the synthesis occurs in feature or image space. \framework surpasses the feature-space methods VOS and NPOS by~63.14 and~35.84 on CIFAR100, respectively. The improvement over the other pixel-space method, Dream-OOD, is particularly striking, demonstrating that the performance gains can be attributed to the quality of the generated outliers, as training hyperparameters were kept unchanged. We provide qualitative examples and an in-depth ablation study in the next sections to visualize the faithfulness of our outliers and analyze what drives these gains in performance.

\subsection{Qualitative examples}

\begin{figure}
    \centering
    \begin{tabular}{c}
    \includegraphics[width=\linewidth]{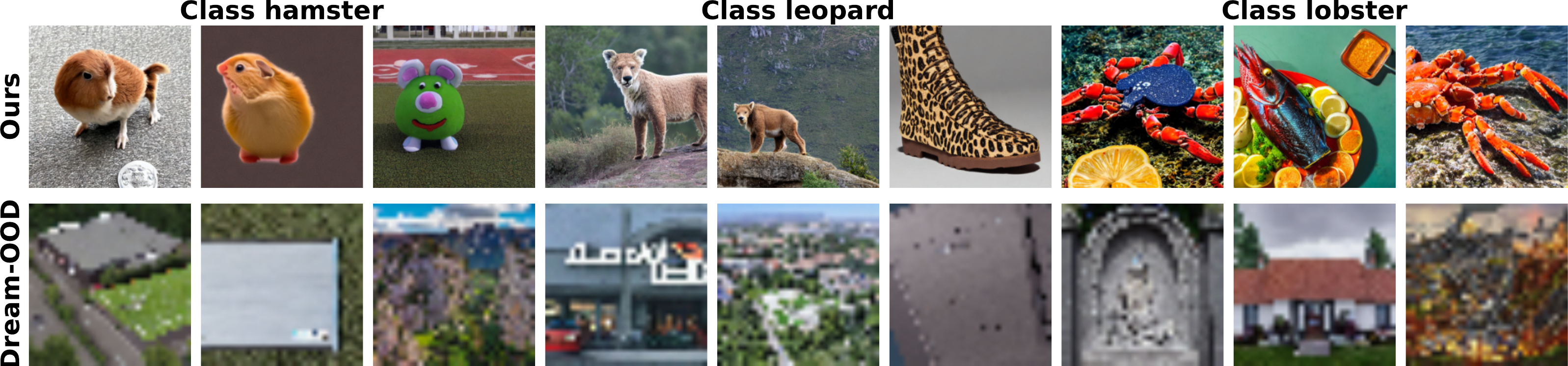} \\
    \includegraphics[width=\linewidth]{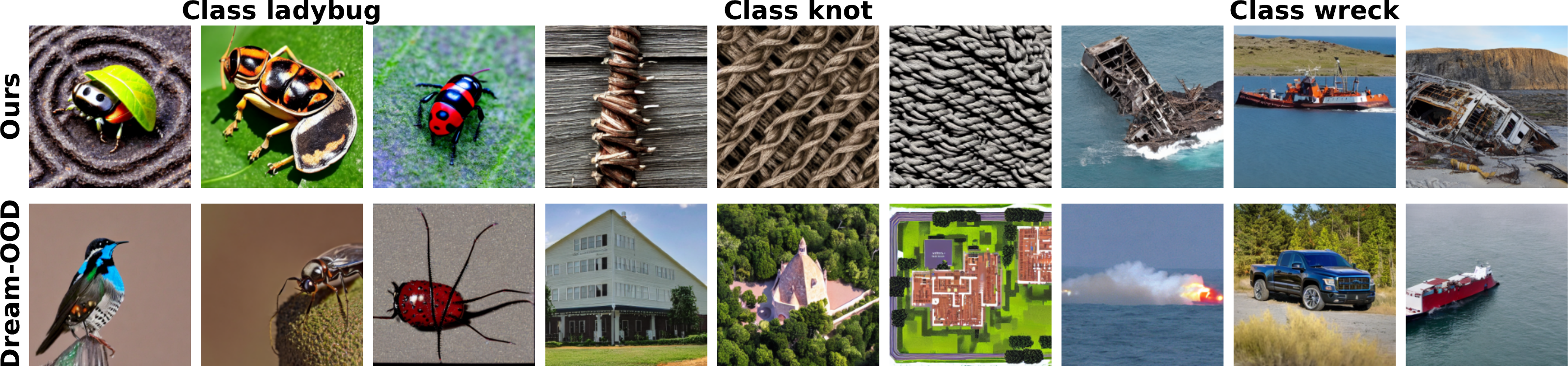} \\
    \end{tabular}
    \caption{\textbf{Nine random generated outliers by our method and Dream-OOD for CIFAR100 (top) and ImageNet-100 (bottom).} Our generated outliers are closer to the intended meaning, providing better signal to learn the ID/OOD decision boundary.}
    \label{fig:quali}
\end{figure}

Fig.~\ref{fig:quali} shows qualitative examples of outliers generated by our method along with images generated by Dream-OOD, taken directly from the official repository\footnote{\url{github.com/deeplearning-wisc/dream-ood}}. For CIFAR100, these images are provided in 32x32, explaining the lower resolution. The Dream-OOD images bear little resemblance to their supposed class. In contrast, ours are closer to the original meaning but still clearly show outlying attributes, such as a leopard-print boot instead of a leopard or a green clay hamster. Our near-OOD samples provide a better signal for OOD detection.

\subsection{Ablation studies}

\begin{table}[t]
  \centering
  \small
    \begin{tabular}{ccccccc}
    \toprule
     Resizing & Sampling & Embeddings & Mean FPR \\
    \midrule
  Nearest & NPOS & Dream & 48.8 \\
  Bilinear & NPOS	& Dream	&	33.5 \\
  Bilinear & Linear invariants & Dream & 29.5 \\
  Bilinear & Linear invariants & Ours & 20.2 \\
  Bilinear & cVPN & Ours & 18.6 \\
  Both & cVPN & Ours & 12.8 \\
   \hline
\end{tabular}
 \caption{\textbf{Ablating \framework on CIFAR100.} We report the mean FPR95 over the five OOD datasets, training with 6400 outliers. All components are important to reach the best performance.}
    \label{tab:abl}
\end{table}

\begin{figure}[t]
  \centering
  \setlength\tabcolsep{1pt}
  \begin{tabular}{cc}
    \includegraphics[width=0.45\linewidth]{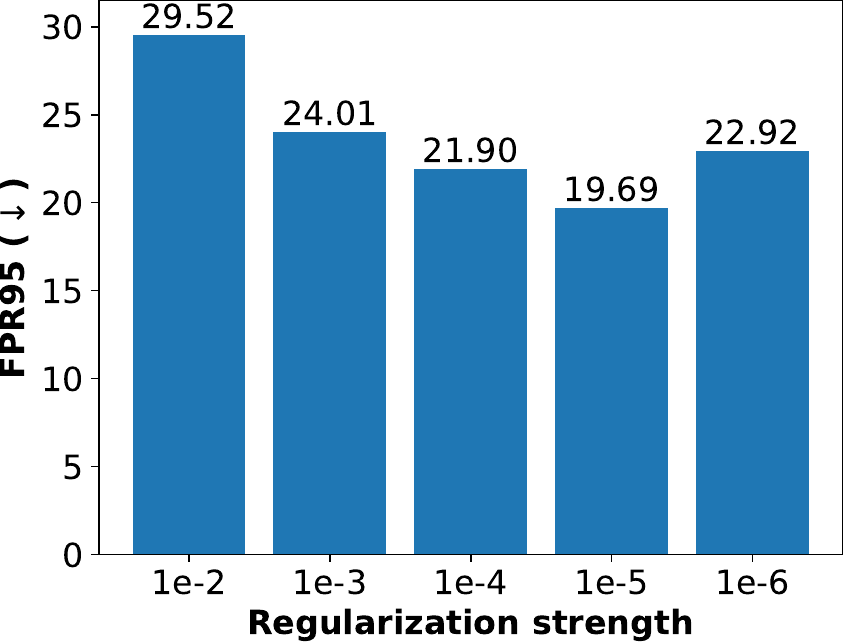} &
    \includegraphics[width=0.45\linewidth]{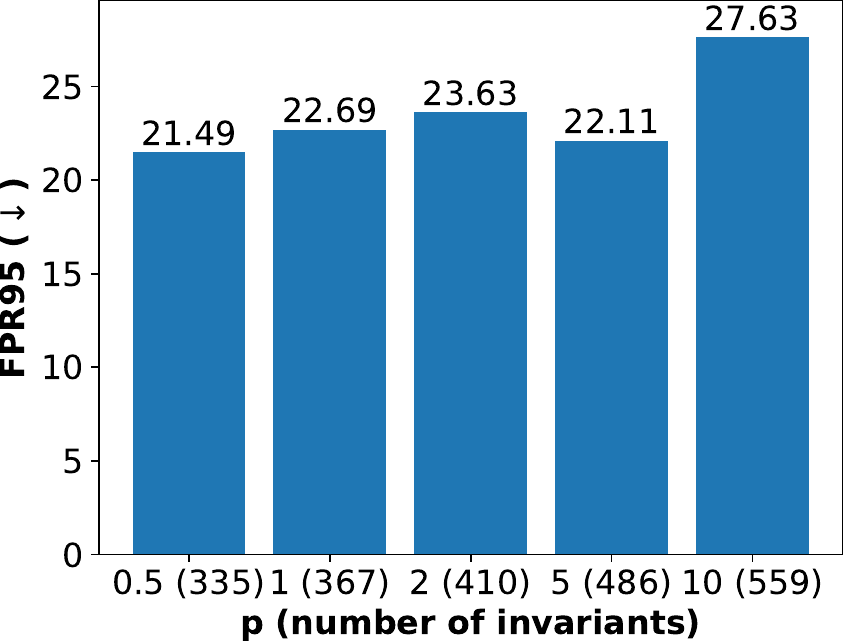} \\
    (a) & (b) \\
  \end{tabular}
    \caption{\textbf{Effect of $\lambda$ (a) and $p$ (b) on the FPR95 ($\downarrow$) with CIFAR100 as the in-distribution.} Difficult samples (see Fig.~\ref{fig:reg}) lead to better OOD detection until they become too similar to the training data. \framework is also robust with respect to the number of invariants up to around $p=10$.}
    \label{fig:reg_abl}
\end{figure}

We ablated the impact of three specific components in \framework on CIFAR-100 to assess their contributions:
\begin{description}
    \item[Embedding type.] We compare our diffusion space embeddings (Sec.~\ref{sec:embedding}) against Dream-OOD's embeddings.
    \item[Sampling distributions.] We evaluate our non-linear probability distributions modeled via the cVPN (Sec.~\ref{sec:nonlinear}) against simpler Gaussian distributions fitted to the ID embeddings in diffusion space and against the sampling method from NPOS.
    \item[Pixel interpolation method.] Synthetic OOD samples are resized before being used to compute the regularization term (Sec.~\ref{sec:regularization}). We compare bilinear and nearest neighbor interpolation, as well as randomly applying bilinear or nearest neighbor interpolation during training as data augmentation. In all cases, the interpolation method used at test time is fixed to match the evaluation protocol used in~\cite{du2024dream}.
\end{description}
Our findings indicate that the three components are essential to achieve optimal performance (Fig.~\ref{fig:reg_abl}). We found that the pixel interpolation method had a more pronounced effect than anticipated. In particular, we observed large drops in performance that seemed to appear when the interpolation methods used at training and testing were mismatched. This finding suggests that OOD detectors focus more on low-level statistics than semantic features. Similar conclusions have been found in other OOD subfields~\cite{havtorn2021hierarchical,schirrmeister2020understanding,serra2019input}, but this is the first demonstration of this effect in the outlier synthesis context. Note that our method still outperforms Dream-OOD when using the same resizing strategy, and all other baselines are unaffected by the interpolation method. Results in Tab.~\ref{tab:abl} also suggest that performing data augmentation on the interpolation method at training time is the most effective approach to dealing with this issue.

\textbf{Hyperparameter sensitivity.} 
We investigate the impact of the regularization factor~$\lambda$ and the hyperparameter controlling the number of invariants~$p$ on OOD detection performance using classifiers trained with 3200~generated outliers (Fig.~\ref{fig:reg_abl}). We observe that lower values of~$\lambda$, which produce samples closer to the training data manifold, generally improve OOD detection up to a threshold around~$\lambda=10^{-6}$, after which performance begins to decline.
Similarly, our method is robust to variations in~$p$, with substantial changes in performance arising only at extreme values. 

\textbf{Different architectures.}
We compare the ViT-B/16 and ConvNeXt-B architectures on ImageNet-100 in Tab.~\ref{tab:arch}. Larger architectures, which achieve better classification performance, also improve OOD detection when used with \framework. This result highlights the flexibility of our method, which performs well across different architectures.

\begin{table}[t]
  \centering
  \small
  \resizebox{\linewidth}{!}{
    \begin{tabular}{ccccccc}
    \toprule
    Methods  & \textsc{iNat} & \textsc{Places} & \textsc{Sun} & \textsc{Tex} & Mean & Acc  \\
    \midrule
 \textit{ResNet-34}  & 20.1 & 	35.8 & 	35 & 	48.5 & 34.9 & 87.1  \\
 \textit{ViT-B/16}  & 18.0 & 		30.0 & 	36.6 & 	18.7 & 25.8 &	92.9 \\
 \textit{ConvNeXt} & 12.4 & 	32.2 & 	 	25.9 & 	18.1 & 	22.2 & 94.3  \\
   \hline
\end{tabular}
}
 \caption{\textbf{Results in FPR95 using different architectures on ImageNet100.} \framework is successful with large vision transformers and modern CNNs.}
    \label{tab:arch}
\end{table}

\begin{figure}[t]
  \centering
  \setlength\tabcolsep{1pt}
  \begin{tabular}{cc}
    \includegraphics[width=0.45\linewidth]{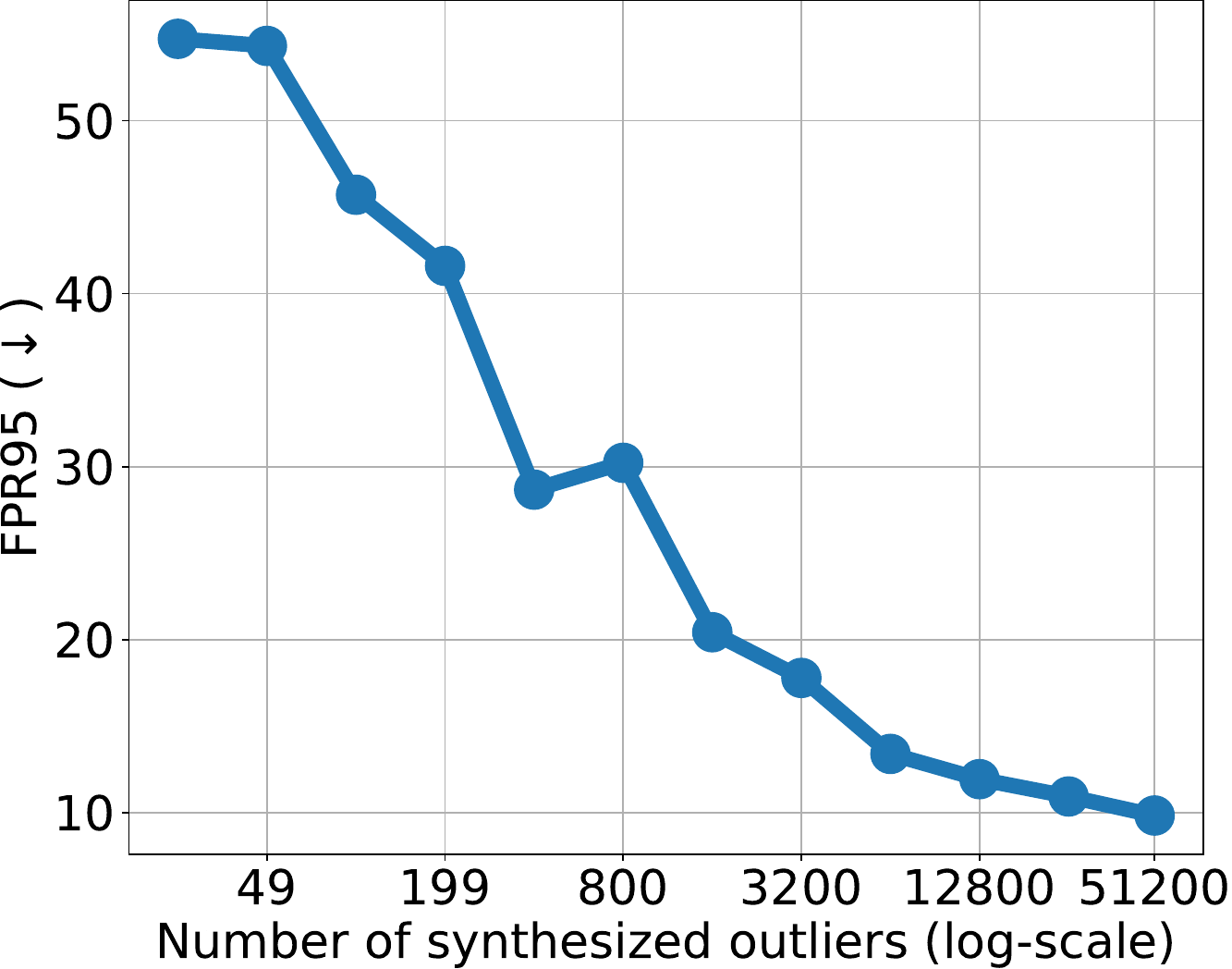} &
    \includegraphics[width=0.45\linewidth]{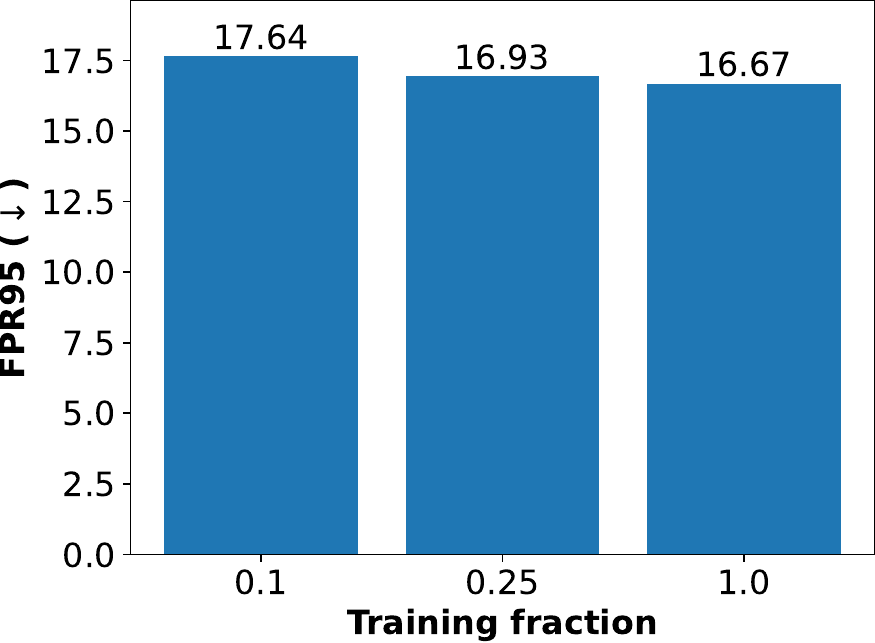} \\
    (a) & (b) \\
  \end{tabular}
    \caption{\textbf{Effect of (a) the number of synthetic outliers (log-scale) and (b) the number of training samples on the OOD detection performance}. (a) shows that a large synthetic dataset is important to achieve good performance. (b) shows that \framework can safely use a fraction of the training data without sacrificing much performance (results with 3200 outliers).}
    \label{fig:number}
\end{figure}

\textbf{Number of outliers.}
We examine how the quantity of synthetic outliers used during training impacts the performance in Fig.~\ref{fig:number}(a). We observe a power-law relationship where increasing the number of outliers significantly enhances OOD detection. This suggests that generating a large synthetic dataset is key to achieving good results. 

\textbf{Computational cost.}
Compared to~\cite{du2024dream}, our approach simplifies training by removing the need for training a text-conditioned latent space and sampling from its embeddings. Instead, we introduce the need to train the cVPN and obtain the diffusion embeddings. The overhead the cVPN introduces is minimal, as training takes 30-40 minutes on a single RTX3090 GPU, and the sampling cost is negligible. However, obtaining the embeddings of the training data is computationally more intensive, taking around 13~hours using 8~A100 GPUs for CIFAR100.
Nonetheless, results in Fig.~\ref{fig:number}(b) suggest that this computational load can be reduced with minimal impact on performance by limiting the number of ID embeddings used for fitting ID probability distributions. 
\section{Conclusion}

This work presents \framework, a novel method for generating outliers to enhance OOD detection. By operating in the embedding space of Stable Diffusion and modeling complex distributions within the training data through a conditional volume-preserving network (cVPN), \framework improves upon prior methods to achieve state-of-the-art results on two widely used OOD benchmarks. Through comprehensive ablation studies, we assess the importance of each component in our approach. Our findings also highlight that OOD detectors are easily misled by low-level image statistics rather than image semantics, underscoring the need for careful treatment of such features in future designs. Overall, we show that outlier synthesis effectively boosts OOD detectors without needing labor-intensive collection and curation of real OOD samples. 

A current limitation of \framework{} is its reliance on the frozen image decoder of Stable Diffusion, which constrains its ability to generate realistic outliers for domains like medical imaging. Future work will explore the benefits of using domain-specific diffusion models for outlier synthesis.

{
    \small
    \bibliographystyle{ieeenat_fullname}
    \bibliography{main}
}

\clearpage
\setcounter{page}{1}
\maketitlesupplementary

\section{Dataset Details}

We follow the exact experimental protocol of~\cite{du2024dream}. The ID datasets are CIFAR-100 and ImageNet-100, which we briefly describe below:
\begin{description}
 \item[CIFAR-100~\cite{krizhevsky2009learning}] contains 50'000 training images and 10'000 testing images belonging to 100 classes. 
 \item[ImageNet-100] is a subset of the full ImageNet~\cite{deng2009imagenet} dataset. We take the 100 classes sampled by~\cite{du2024dream} for a total of 129'860 training samples and 5'000 test samples. These classes are: {\scriptsize n01498041, n01514859, n01582220, n01608432, n01616318,
        n01687978, n01776313, n01806567, n01833805, n01882714,
          n01910747, n01944390, n01985128, n02007558, n02071294,
          n02085620, n02114855, n02123045, n02128385, n02129165,
          n02129604, n02165456, n02190166, n02219486, n02226429,
          n02279972, n02317335, n02326432, n02342885, n02363005,
          n02391049, n02395406, n02403003, n02422699, n02442845,
          n02444819, n02480855, n02510455, n02640242, n02672831,
          n02687172, n02701002, n02730930, n02769748, n02782093,
          n02787622, n02793495, n02799071, n02802426, n02814860,
          n02840245, n02906734, n02948072, n02980441, n02999410,
          n03014705, n03028079, n03032252, n03125729, n03160309,
          n03179701, n03220513, n03249569, n03291819, n03384352,
          n03388043, n03450230, n03481172, n03594734, n03594945,
          n03627232, n03642806, n03649909, n03661043, n03676483,
          n03724870, n03733281, n03759954, n03761084, n03773504,
          n03804744, n03916031, n03938244, n04004767, n04026417,
          n04090263, n04133789, n04153751, n04296562, n04330267,
          n04371774, n04404412, n04465501, n04485082, n04507155,
          n04536866, n04579432, n04606251, n07714990, n07745940}.

\end{description}

For CIFAR-100, we use the test sets of five different datasets as OOD:
\begin{description}
    \item[SVHN~\cite{netzer2011reading}] containing 10'000 images of house numbers.
    \item[Places365~\cite{zhou2017places}] is a dataset of large scenes, where we use the 10'000 random images sampled and provided by~\cite{sun2022out,tao2023nonparametric}.
    \item[Lsun~\cite{yu2015lsun}] is a large-scale dataset of scenes and objects. We use the subset of 10'000 images provided by~\cite{sun2022out,tao2023nonparametric}.
    \item[iSun~\cite{xu2015isun}] contains images of natural scenes, where we use the subset of 10'000 images provided by~\cite{sun2022out,tao2023nonparametric}.
    \item[Textures~\cite{cimpoi2014describing}] has 5'640 images of patterns and textures.
\end{description}

For ImageNet-100, we use four datasets where the classes of the test sets do not overlap with the full Imagenet, as provided by~\cite{huang2021mos}:
\begin{description}
\item[iNaturalist~\cite{van2018inaturalist}] has images of plants and animals. We use a 10'000 image subset of 110 plant classes not present in ImageNet~\cite{du2024dream}.
\item[SUN~\cite{xiao2010sun}] contains images of natural scenes, where we use a 10'000 image subset of 50 natural objects not present in Imagenet~\cite{du2024dream}.
\item[Places~\cite{zhou2017places}] is a dataset of large scenes, where we use 10'000 images from 50 categories that are not present in Imagenet~\cite{du2024dream}.
\item[Textures~\cite{cimpoi2014describing}] has 5'640 images of patterns and textures.

\end{description}


\end{document}